\documentclass[conference, doublecolumn, hidelinks, xcolor=dvipsnames]{IEEEtran}

\usepackage{microtype}
\usepackage{acronym}
\usepackage{amsmath}
\usepackage{amssymb}
\usepackage[english]{babel}
\usepackage{bbm}
\usepackage[shortlabels,inline]{enumitem}
\usepackage{graphicx}
\usepackage{hyperref}
\usepackage{cleveref}
\usepackage[utf8]{inputenc}
\usepackage{lipsum}
\usepackage{siunitx}
\usepackage[dvipsnames]{xcolor}
\usepackage[T1]{fontenc}
\usepackage{threeparttable}
\usepackage{csquotes}
\usepackage{multicol}
\usepackage{booktabs}
\usepackage{cite}

\allowdisplaybreaks

\newtheorem{corollary}{Corollary}

\newtheorem{definition}{Definition}

\newcommand{\Section}[1]{Section~\ref{#1}}

\DeclareMathOperator*{\argmax}{arg\,max}

\newcommand{\setF}{\ensuremath{\mathcal{F}}}
\newcommand{\setS}{\ensuremath{\mathcal{S}}}
\newcommand{\setXs}{\ensuremath{\mathcal{X}^{\mathrm{s}}}}
\newcommand{\setXc}{\mathcal{X}^{\mathrm{c}}}
\newcommand{\subsetF}{\ensuremath{\mathbf{F}}}
\newcommand{\subsetS}{\ensuremath{\mathbf{S}}}
\newcommand{\subsetXs}{\mathbf{X}^{\mathrm{s}}}
\newcommand{\subsetXc}{\mathbf{X}^{\mathrm{c}}}

\acrodef{AD}{Alzheimer's disease}
\acrodef{ASD}{autism spectrum disorder}
\acrodef{CNS}{central nervous system}
\acrodef{DRE}{drug refractory epi\-lep\-sy}
\acrodef{GBA}{gut-brain axis}
\acrodef{GBA-MC}{GBA-based MC}
\acrodef{HFD}{high-fat diet}
\acrodef{HK}{high-fat to ketogenic conversion}
\acrodef{IoBNT}{Internet of Bio-Nano-Things}
\acrodef{KD}{ketogenic diet}
\acrodef{LCD}{low-carbohydrate diet}
\acrodef{MC}{molecular communication}
\acrodef{MI}{mutual information}
\acrodef{MS}{multiple sclerosis}
\acrodef{ND}{normal diet}
\acrodef{RF}{random forest}
\acrodef{RV}{random variable}
\acrodef{Rx}{receiver}
\acrodef{SMC}{synthetic molecular communication}
\acrodef{SM}{signaling molecule}
\acrodef{Tx}{transmitter}
\acrodef{wrt}[w.r.t.]{with respect to}

\begin{document}
\bstctlcite{IEEEexample:BSTcontrol}

\title{Synthetic MC via Biological Transmitters:\\ Therapeutic Modulation of the Gut-Brain Axis}

\author{
\IEEEauthorblockN{Sebastian Lotter\IEEEauthorrefmark{1}, Elisabeth Mohr\IEEEauthorrefmark{1}, Andrina Rutsch\IEEEauthorrefmark{2}, Lukas Brand\IEEEauthorrefmark{1}, Francesca Ronchi\IEEEauthorrefmark{2}, Laura Díaz-Marugán\IEEEauthorrefmark{2}}\\[-0.2cm]
\IEEEauthorblockA{\small\IEEEauthorrefmark{1}Friedrich-Alexander-Universität Erlangen-Nürnberg, Erlangen, Germany}\\
\vspace{-0.35cm}
\IEEEauthorblockA{\small\IEEEauthorrefmark{2}Charité – Universitätsmedizin Berlin, Humboldt-Universität zu Berlin, Berlin Institute of Health (BIH), Berlin, Germany}\\
\vspace{-1.3cm}
}

\maketitle
\thispagestyle{plain}
\pagestyle{plain}
\begin{abstract}
\Ac{SMC} is a key enabler for future healthcare systems in which \ac{IoBNT} devices facilitate the continuous monitoring of a patient's biochemical signals.
To close the loop between sensing and actuation in these systems, both the {\em detection} and the {\em generation} of in-body \ac{MC} signals is key.
However, generating signals inside the human body, e.g., via synthetic nanodevices, still poses a major research challenge in \ac{SMC}, due to technological obstacles as well as legal, safety, and ethical issues.
In contrast to many existing studies, this paper considers an \ac{SMC} system in which signals are generated {\em indirectly} via the modulation of a natural in-body \ac{MC} system, namely the \ac{GBA}.
Therapeutic \ac{GBA} modulation is already established as treatment for some neurological diseases, e.g., \ac{DRE}, and performed via the administration of nutritional supplements or specific diets with therapeutic effect.
However, the molecular signaling pathways that mediate the effect of such treatments are mostly unknown.
Consequently, existing treatments are standardized or designed heuristically and able to help only some patients while failing to help others.
In this paper, we propose to leverage personal health data, e.g., data gathered by in-body \ac{IoBNT} devices, to overcome this research gap and design more versatile and robust \ac{GBA} modulation-based treatments as compared to the existing ones.
To show the feasibility of our approach, we first define a catalog of theoretical requirements for therapeutic \ac{GBA} modulation.
Then, we propose a machine learning model to verify these requirements for practical scenarios when only limited data on the \ac{GBA} modulation is available.
By evaluating the proposed model on several published datasets, we confirm its excellent accuracy in identifying different modulators of the \ac{GBA}.
Finally, we utilize the proposed model to identify specific modulatory pathways that play an important role for therapeutic \ac{GBA} modulation.
The results presented in this paper may help to develop novel personalized \ac{GBA}-based treatments, i.e., novel nutritional supplements and/or diets, to help patients that do not respond to existing standardized treatments.

\end{abstract}

\setlength{\belowdisplayskip}{2pt}
\setlength{\belowdisplayshortskip}{2pt}
\acresetall
\section{Introduction}\label{sec:intro}
\Ac{SMC} is expected to facilitate early disease detection and treatment with the help of synthetic in-body nanodevices that operate in the {\em \ac{IoBNT}} \cite{akyildiz2015internet}.
The \ac{IoBNT} is a next-generation communication network that integrates conventional communication networks, such as the Internet, with nanoscale in-body communication devices, like nanorobots and synthetic cells.
Conceptually, many of the disruptive healthcare applications enabled by \ac{SMC} involve two main components: {\em sensing} and {\em actuation}.
Here, sensing refers to the detection of natural molecular signals inside the human body.
For example, biomarkers indicative of some disease can be sensed.
The detected natural \ac{MC} signals can then be processed and interpreted either locally, i.e., in the body, or remotely, i.e., after relaying them to computing devices outside the body and/or human specialists.
If necessary, this can trigger personalized medical interventions.

In turn, actuation entails generating \ac{MC} signals inside the human body to enable specific, targeted medical interventions.
This signal generation can happen {\em directly} or {\em indirectly}.
In direct signal generation, \ac{MC} signals are generated by {\em engineered devices}, e.g., in-body nanorobots or synthetic cells, whereas {\em biological signal generators} are used to generate the \ac{MC} signals in indirect signal generation.
The direct generation of molecular signals for therapeutic applications has widely been researched in the \ac{MC} literature in theory, especially in the context of targeted drug delivery and controlled release \cite{chude2017molecular, chahibi2013molecular}.
However, the existing administrative and legal requirements render the admission of active (i.e., signal generating) in-body nanodevices for medical applications a challenging and potentially long-term process in practice.
In contrast, this paper considers an \ac{SMC} system in which the communication signals are generated indirectly, namely the \ac{GBA}.

\begin{figure}
    \centering
    \includegraphics[width=0.75\linewidth]{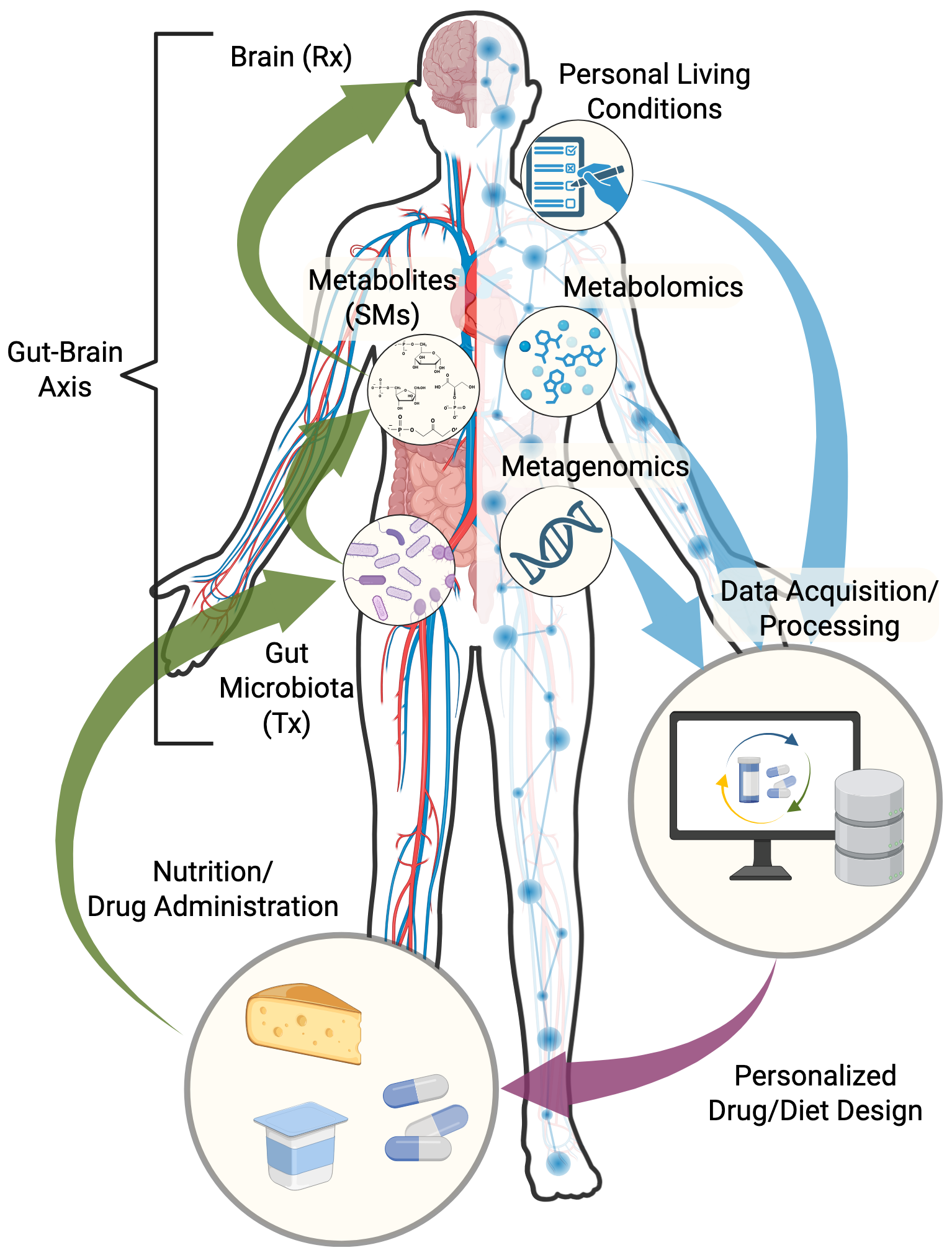}
    \caption{\small The integration of next-generation health data acquisition facilitated by the \ac{IoBNT} (right, blue arrows) with therapeutical modulation of the \ac{GBA} (left, green arrows) via personalized, data-informed drug/diet design (bottom, purple arrow). In \ac{GBA-MC}, the gut microbiota acts as \ac{MC} \ac{Tx} producing metabolites as \acp{SM} that propagate through blood vessels and the blood-brain-barrier to the brain acting as the \ac{MC} \ac{Rx}.}
    \label{fig:overview}
    \vspace*{-2mm}
\end{figure}

The \ac{GBA} has been the subject of several studies in physiological and pathological conditions \cite{dinan2017microbiome} and has already been identified as one promising research direction for \ac{SMC} \cite{akyildiz2019microbiome,somathilaka2022graphbased}.
Also, modulation of the \ac{GBA} via dietary interventions and nutritional supplements is an established treatment for neurological disorders such as \ac{DRE} \cite{effinger2023ketogenic,iannone2019microbiota,lum2023ketogenic}.
However, it is currently not known how such modulation affects the generation of \ac{MC} signals along the \ac{GBA}, limiting the possibilities to adapt the treatment to the individual patient's biochemical preconditions (e.g., their gut microbiome composition).
Overcoming this lack of knowledge will facilitate the development of more robust and more effective treatments as compared to existing ones.
In this paper, we showcase how personalized health data, such as that acquired by future \ac{IoBNT} devices, can be used to understand the interaction between medical interventions and \ac{MC} along the \ac{GBA} (\acsu{GBA-MC}).
Specifically, we utilize a novel information theoretical framework for \ac{GBA-MC} to develop a computational tool that identifies main modulatory effects of dietary interventions on \ac{GBA-MC} and confirm the accuracy of this tool with data from mice models.
Specifically, this paper makes the following main contributions:
\begin{itemize}
    \item An information theoretical framework for the indirect signal generation via modulation of a biological \ac{MC} \ac{Tx} is proposed for an \ac{MC} system that is highly relevant for medical applications, namely the \ac{GBA}.
    \item Based on the information theoretical framework, a novel computational model for the dietary modulation of \ac{GBA-MC} based on a \ac{RF} classifier is presented.
    \item The proposed model is utilized to study the modulation of the \ac{GBA} by specific dietary interventions under various conditions (e.g., health status) of the host.
    \item Main modulatory pathways are identified \ac{wrt} the gut bacteria and the associated metabolic signaling.
\end{itemize}
In summary, this paper presents a step towards closing the sensing-actuation loop in the \ac{IoBNT}-assisted treatment of neurological disorders (as illustrated in Fig.~\ref{fig:overview}).
Due to the generality of the proposed framework, it can be adapted to the modulation of other biological \ac{MC} systems.

The remainder of this paper is organized as follows.
\Section{sec:background} introduces the relevant biological background on the \ac{GBA} and its importance for the dietary treatment of neurodegenerative diseases.
\Section{sec:system_model} introduces the novel information theoretical framework for \ac{GBA-MC}.
The proposed computational model is presented in \Section{sec:random_forest_model} and the utilized datasets are introduced in \Section{sec:data_sources}.
Finally, \Section{sec:results} presents the main evaluation results and \Section{sec:conclusion} concludes the paper and outlines topics for future~work.

%

\section{Gut-Brain Axis, Dietary Interventions, and Neurodegenerative Disease}\label{sec:background}
\subsection{Background}

The \ac{GBA} is the multidirectional communication\footnote{We focus on the communication from gut to brain in this paper, hence, only this direction is illustrated in Fig.~\ref{fig:overview}.} between the intestine with commensal microbes (gut microbiota) and the \ac{CNS} with a profound influence on neurological disorders and neural development.
The gut microbiota is a community of symbiotic microorganisms that reach a density of more than $10^{12}$ cells/g of content in the large intestine of mammals \cite{macpherson2004interactions}. 
Under healthy conditions, the mucosal microbiota plays a vital role in food digestion, vitamin synthesis, physiological organ development, and protection against pathogens.
Intestinal microbes have also been shown to be important in the \ac{CNS} development.
Alterations in the balance of host-microbial interactions in the gut, such as dysbiotic (altered in composition) microbiota, have been observed in various neurological disorders, such as major depressive and mood disorders, neurodegenerative, and neuropsychiatric disorders \cite{singh2020global,sirven2015epilepsy}.

Mounting evidence points towards gut microbiota as an essential mediator of dietary effects on the host organism \cite{david2014diet}.
Intestinal microbiota can affect the host responses through the dissemination of bacterial products or metabolites (substances produced during metabolism) and this is strongly dependent upon the substrates available through dietary intake \cite{devadder2014microbiota}. 
Diet is, in fact, one of the major forces that shape the microbiota \cite{zhernakova2016population,turnbaugh2009effect,falony2016population} in terms of composition and metabolism, contributing to host homeostasis and disease susceptibility.
Currently, nutrition is a complementary and alternative approach in the management of neurological diseases, showing positive effects, especially in drug-resistant patients \cite{diazmarugan2023microbiota,ceccon2024personalized,iannone2019microbiota}.

A dietary approach that has been shown to ameliorate several metabolic and neurological conditions, such as \ac{DRE}, \acl{ASD}, \acl{AD}, and cancer is a {\em \ac{KD}}.
\ac{KD}, high in fat and low in carbohydrates, decreases glycemia and insulin levels, and increases the concentration of ketone bodies systemically.
Recently, clinical and experimental studies suggested important effects of \ac{KD} regimen on the host \ac{CNS} via alterations of the microbiota.
We will conceptualize these modulatory effects of \ac{KD} on the \ac{GBA} information theoretically in \Section{sec:system_model}.
To exemplify first some major open research questions associated with \ac{KD}-based treatment, we focus on the treatment of \ac{DRE} in the following.
\ac{DRE} accounts for one third of the people with epilepsy and 80\% of the total cost of treating the condition.
People with \ac{DRE} have been reported to have an enrichment of several microbes (e.g., {\em Clostridium XVIII}, {\em Dorea}, {\em  Coprobacillus}, {\em Ruminococcus}, {\em Akkermansia}, {\em Neisseria}, {\em Coprococcus} or {\em Actinobacteria}, {\em Verrucomicrobia}, {\em Nitrospirae}, {\em Blautia}, {\em Bifidobacterium}, {\em Subdoligranulum}, {\em Dialister}, and {\em Anaerostipes} \cite{peng2018altered,gong2020alteration}), compared to drug-sensitive patients and healthy controls.
In a study of $20$ people affected by \ac{DRE}, after six months of \ac{KD} treatment, a quarter of the people had $\geq 90\%$ or complete seizure reduction, another quarter had seizure reduction of $50$-$89\%$, and half had $< 50\%$ reduction \cite{zhang2018altered}.
Indeed, treatment with \ac{KD} can reduce seizure frequency by about $50\%$ in about $50\%$ of \ac{DRE} cases that adhere to it.
Animal and human studies suggest that gut microbiota is responsible for the benefits of this diet \cite{olson2018gut,lum2023ketogenic}.
Furthermore, Dahlin {\em et al.}~identified associations between specific microbial species and serum metabolites associated with seizure reduction in people with \ac{DRE} after \ac{KD} intervention \cite{dahlin2024association}.
These data show the potential and the importance of deciphering how the diet and microbiota can modulate disease development in some individuals with beneficial or detrimental effects on the \ac{CNS} health.
However, causal relationships between the composition of the human gut microbiota and the therapeutic effect of \ac{KD} remain elusive.
Elucidating such relationships could help to predict and further improve the success rate of \ac{KD}-based treatment and will pave the way for future therapies for various disorders. This is especially important since \ac{KD} is a very restrictive diet with low acceptance or adherence to, and its long-term consumption could cause some adverse effects, such as nausea, vomiting, diarrhea, constipation, and weight loss.

\subsection{Integration with the IoBNT}

\ac{KD}, as a specific therapeutic measure in the treatment of \ac{DRE}, exemplifies (i) the enormous benefit for public health that would result from improving the efficiency of \ac{GBA}-mediated therapies and (ii) that such improvements could possibly be achieved by better aligning the treatments with the individual gut microbiota composition of the host.
Now, the \ac{IoBNT}, in which the host's body functions are continuously monitored at unprecedented levels, enables access to metagenomic and metabolomic data that characterizes the state of the host's gut microbiome and its response to dietary interventions.
Facilitated by this data (and by additional information on the host's living conditions, e.g., live habits, psychosocial factors), personalized treatments that are aligned with the host's gut microbiota come in reach.
Fig.~\ref{fig:overview} illustrates this concept: data acquired by \ac{IoBNT} devices is processed at central computing units and utilized to design personalized diets or nutritional supplements.
When administered, such personalized treatments are not only expected to be more effective in the management of neurological diseases as compared to existing standardized treatments, their effect can also be sensed directly by the \ac{IoBNT} devices which closes the loop by facilitating further adjustments of the treatment.
Next, we present a feasibility study that demonstrates how the \ac{SMC} framework can help to realize this vision.

\section{System Model and Problem Formulation}\label{sec:system_model}
One major challenge in designing personalized dietary treatments or nutritional supplements for combating neurodegenerative diseases via the \ac{GBA} is to identify the main signaling pathways by which externally controllable variables, such as the diet, impact communication along the \ac{GBA}.
In the following, we first introduce a formal framework for gut microbiota-mediated signaling, before we formulate the inference problem that we will utilize later on to tackle this problem.

\subsection{System Model}

We model the concentrations of different bacterial species $i \in \mathcal{B}$ as \acp{RV} $S_i \in \mathbb{R}_0^{+}$, where $\mathcal{B}$ and $\mathbb{R}_0^{+}$ denote the set of bacterial species that are potentially present in the gut and the set of non-negative real numbers, respectively.
In \ac{GBA-MC}, the gut microbiome is regarded as \ac{Tx} with associated state $\mathcal{S} = \lbrace S_i \rbrace$ that emits molecular signals composed of different metabolites (corresponding to \acp{SM}).
Depending on $\mathcal{S}$, the {\em metabolic profiles}, i.e., the presence and absence of particular metabolites, of other parts of the human body vary.
In \ac{GBA-MC}, we are mostly interested in the metabolic profile of the serum, which we denote by $\mathcal{X}^{\mathrm{s}} = \lbrace X^{\mathrm{s}}_k \rbrace$, where $k \in \mathcal{M}$, $\mathcal{M}$ denotes the set of metabolites, and \ac{RV} $X^{\mathrm{s}}_k \in \mathbb{R}_0^+$ denotes the concentration of metabolite $k$ in the serum.
Via the serum, in particular via $\mathcal{X}^{\mathrm{s}}$, information is conveyed from the gut to the brain, cf.~\Section{sec:background}.
However, $\mathcal{X}^{\mathrm{s}}$ is subject to many other influences besides $\mathcal{S}$, such as the health status of the host.
Hence, the statistical correlation between $\mathcal{X}^{\mathrm{s}}$ and $\mathcal{S}$ may be obfuscated by interfering molecular signals, e.g., signals originating from cancerous tissue.
Hence, as an additional readout for the metabolic activity of the gut microbiota, we consider in this paper also the metabolic profile of the colon, denoted as $\mathcal{X}^{\mathrm{c}} = \lbrace X^{\mathrm{c}}_k \rbrace$, where \ac{RV} $X^{\mathrm{c}}_k \in \mathbb{R}_0^+$ denotes the concentration of metabolite $k$ in the colon.
$\mathcal{X}^{\mathrm{c}}$ comprises the molecular signal released from the gut microbiota to the colon.
Due to the spatial proximity between gut microbiota and colon metabolites, $\mathcal{S}$ and $\mathcal{X}^{\mathrm{c}}$ are statistically strongly correlated.

Both, the gut microbiota itself ($\mathcal{S}$) and the metabolic signals modulated by it ($\mathcal{X}^{\mathrm{s}}$ and $\mathcal{X}^{\mathrm{c}}$) are subject to multifactorial dependencies, among others, on the genetic identity, current constitution, and alimentation/diet of the host, cf.~\Section{sec:background}.
In this paper, we characterize some of these dependencies using previously published data from mice and statistical inference with the long-term goal to identify novel targets for therapeutic modulation of the \ac{GBA} in the personalized treatment of neurodegnerative diseases.

To introduce the proposed approach rigorously in the following, we finally collect factors that are potentially statistically interdependent with $\mathcal{S}$, $\mathcal{X}^{\mathrm{s}}$, or $\mathcal{X}^{\mathrm{c}}$ in the set of random variables $\mathcal{F} = \lbrace F_j \rbrace$, $j \in \mathbb{N}$.
In the context of this paper, $F_j$ are discrete-valued \acp{RV} indicating the value of any categorical variable.
For example, $F_j$ can denote the type of diet on which the host is fed or the type of disease (e.g., cancer type) it is infected with.
However, the framework presented in this paper can easily be extended to continuous-valued $F_j$'s.
The statistical interdependencies between $\mathcal{F}$ on the one hand and $\mathcal{S}$, $\mathcal{X}^c$, and $\mathcal{X}^s$ on the other hand are unknown in general and object of this study.
Finally, we note that $\mathcal{F}$, $\mathcal{S}$, $\mathcal{X}^c$, and $\mathcal{X}^s$ are distributed according to {\em population statistics}.
Specifically, the randomness in $\mathcal{F}$, $\mathcal{S}$, $\mathcal{X}^c$, and $\mathcal{X}^s$ represents the variability of the respective variables over a {\em population of hosts} (e.g., a group of lab mice), {\em not} the statistics of their potential fluctuations over time in a single host.
On a population level, time (e.g., the time elapsed since the start of a treatment/diet) is accounted for in the proposed model as one of the variables collected in $\mathcal{F}$; this is the case, for example, in Dataset 1 (DS-1) as introduced in \Section{sec:data_sources}.

\subsection{Problem Formulation}

From an information theoretical point of view, the information conveyed between two disjoint sets of \acp{RV} is measured in terms of their {\em \ac{MI}}.
Considering the two finite-sized sets $\mathbf{F} = \lbrace F_{j_1}, \ldots, F_{j_{\vert \mathbf{F} \vert}}\rbrace \subseteq \mathcal{F}$ and $\mathbf{S} = \lbrace S_{i_1}, \ldots, S_{i_{\vert \mathbf{S} \vert}} \rbrace \subseteq \mathcal{S}$, for example, where $\vert \mathbf{A} \vert$ denotes the cardinality of set $\mathbf{A}$, the \ac{MI}, $I$, is defined as
\begin{align}
    I(\mathbf{F};\mathbf{S}) &= H(\mathbf{F}) - H(\mathbf{F}\,\vert\,\mathbf{S}) \\\nonumber
    &= H(F_{j_1}, \ldots, F_{j_{\vert \mathbf{F} \vert}}) - H(F_{j_1}, \ldots, F_{j_{\vert \mathbf{F} \vert}} \vert  S_{i_1}, \ldots, S_{i_{\vert \mathbf{S} \vert}}),
\end{align}
where $H(\cdot)$ and $H(\cdot\,\vert\,\cdot)$ denote the entropy function and the conditional entropy function, respectively.
Adopting this perspective, we define the concept of a {\em \ac{GBA} modulator} as follows.
\begin{definition}[GBA modulator]\label{def:gba_modulator}
    Any subset of influencing factors $\mathbf{F} \subseteq \mathcal{F}$ is a {\em modulator of $\mathcal{S}$ (modulator of $\mathcal{X}^s$, modulator of $\mathcal{X}^c$)} if and only if there exists a non-empty, finite-sized subset $\mathbf{S} \subseteq \mathcal{S}$ $(\mathbf{X}^s \subseteq \mathcal{X}^s$, $\mathbf{X}^c \subseteq \mathcal{X}^c)$ for which $I(\mathbf{F};\mathbf{S}) > 0$ $(I(\mathbf{F};\mathbf{X}^s) > 0$, $I(\mathbf{F};\mathbf{X}^c) > 0)$.
    If $\mathbf{F}$ is a modulator of $\mathcal{S}$ or $\mathcal{X}^s$, it is a {\em GBA modulator}.
\end{definition}

Def.~\ref{def:gba_modulator} unifies the two requirements that (i) $\mathbf{F}$ is non-deter\-mi\-ni\-stic ($H(\mathbf{F})>0$), i.e., it varies over the considered population, and (ii) $\mathbf{F}$ is correlated with the set of \acp{RV} it modulates.
However, from a practical point of view, Def.~\ref{def:gba_modulator} does not address two important aspects that are relevant for therapeutic modulation.
For therapeutic modulation, first, the modulator $\mathbf{F}$ as defined in Def.~\ref{def:gba_modulator} needs to be controllable as part of a medical treatment and, second, it shall exhibit robust influence on the \ac{GBA} as some other, non-controllable influencing factors vary.
The following two definitions address these two points.

\begin{definition}[Feasible GBA modulator]\label{def:feasible_modulator}
    Any \ac{GBA} modulator $\mathbf{F}$ is a {\em feasible GBA modulator} if the values of all $F_{j_1}, \ldots, F_{j_{\vert \mathbf{F} \vert}} \in \mathbf{F}$ can be set via human intervention.
\end{definition}

\begin{definition}[Robust GBA modulator]\label{def:robust_modulator}
    Any \ac{GBA} modulator $\mathbf{F}_1 \subseteq \mathcal{F}$ is a {\em robust modulator of $\mathcal{S}$ (robust modulator of $\setXs$, robust modulator of $\setXc$) \ac{wrt} $\mathbf{F}_2 \subseteq \mathcal{F}$} if and only if $I(\mathbf{F}_1;\mathbf{S}|\mathbf{F}_2) = H(\mathbf{F}_1|\mathbf{F}_2) - H(\mathbf{F}_1|\mathbf{F}_2,\mathbf{S}) > 0$ $(I(\mathbf{F}_1;\subsetXs|\mathbf{F}_2) > 0$, $I(\mathbf{F}_1;\subsetXc|\mathbf{F}_2) > 0)$, where $\mathbf{S}$, $\subsetXs$, and $\subsetXc$ were defined in Def.~\ref{def:gba_modulator}.
    If $\mathbf{F}_1$ is a robust modulator of $\mathcal{S}$ or $\setXs$ \ac{wrt} $\mathbf{F}_2$, it is called a {\em robust GBA modulator \ac{wrt} $\mathbf{F}_2$}.
\end{definition}

A specific disease that interacts with the microbiome or the serum metabolites of the host, a cancer for example, can be a GBA modulator according to Def.~\ref{def:gba_modulator}, but not a feasible GBA modulator, cf.~Def.~\ref{def:feasible_modulator}, given that removing the disease is not available as a direct therapeutic option.
Furthermore, for any $\mathbf{F}_1$ that is a robust GBA modulator \ac{wrt} $\mathbf{F}_2$, Def.~\ref{def:robust_modulator} implies $H(\mathbf{F}_1|\mathbf{F}_2) > 0$, i.e., $\mathbf{F}_2$ cannot be a deterministic function of $\mathbf{F}_1$, and (assuming without loss of generality that $\mathbf{F}_1$ modulates $\mathcal{S}$) $H(\mathbf{F}_1|\mathbf{F}_2, \mathbf{S}) < H(\mathbf{F}_1\vert\mathbf{F}_2)$, i.e., $\mathbf{F}_1$, $\mathbf{F}_2$, $\mathbf{S}$ may {\em not} form a Markov chain $\mathbf{F}_1\rightarrow\mathbf{F}_2\rightarrow\mathbf{S}$ or, in other words, $\mathbf{F}_1$ and $\mathbf{S}$ may {\em not} be conditionally independent given $\mathbf{F}_2$.

Now the problem in identifying feasible and robust GBA modulators according to Defs.~\ref{def:gba_modulator}-\ref{def:robust_modulator} is that the conditional probabilities $\Pr\lbrace \subsetF \,\vert\, \subsetS \rbrace$ and $\Pr\lbrace \subsetF \,\vert\, \subsetXs \rbrace$ are unknown and the \ac{MI} is notoriously difficult to estimate from data whenever high-dimensional sets of random variables (such as $\setS$ and $\setXs$) are involved \cite{roulston1999estimating}.
However, we know that $I(\subsetF;\subsetS)$ ($I(\subsetF;\subsetXs)$) is positive if and only if different realizations of $\subsetF$ can be estimated from realizations of $\subsetS$ ($\subsetXs$) more reliably than by randomly guessing.
Adopting this operational perspective, let us consider a particular realization of $\subsetF$, $\mathbf{f}(\mathbf{s})$, under which an observation $\mathbf{s}$ of $\subsetS$ was obtained and the following Bayesian inference problem\footnote{For the sake of brevity and readability, we focus on the modulation of $\setS$ for the rest of this section omitting $\setXs$ and $\setXc$. However, the following discussion, including Corollary~\ref{cor:error} and Def.~\ref{def:empirical_modulator}, applies analogously to $\setXs$ and $\setXc$ instead of $\setS$, if $\subsetS$ is replaced by $\subsetXs$ and $\subsetXc$, respectively.}
\begin{align}
    \hat{\mathbf{f}}(\mathbf{s}) = \max_{\mathbf{f}'} \,\Pr\lbrace \subsetF= \mathbf{f}' \,\vert\, \subsetS=\mathbf{s}  \rbrace,\label{eq:system_model:inference}
\end{align}
where $\mathbf{f}'$ represents any realization of $\subsetF$.
$\hat{\mathbf{f}}(\mathbf{s})$ is the Bayesian optimal (maximum {\em a posteriori}) estimator of $\mathbf{f}(\mathbf{s})$ and it specializes to the following maximum {\em a priori} estimator
\begin{align}
    \bar{\mathbf{f}} = \max_{\mathbf{f}'} \,\Pr\lbrace \subsetF= \mathbf{f}' \rbrace,\label{eq:system_model:priori_estimator}
\end{align}
if and only if $\subsetF$ and $\subsetS$ are statistically independent.
Hence, if the estimation performance of $\hat{\mathbf{f}}$ is better than that of $\bar{\mathbf{f}}$, $\subsetF$ is a modulator of $\setS$.
The following corollary formalizes the necessary condition for $\subsetF$ to be a modulator according to Def.~\ref{def:gba_modulator} that we have just found.
\begin{corollary}\label{cor:error}
    Let $(\mathbf{f}_n,\mathbf{s}_n)$, $n\in\mathbb{N}$, denote a sequence of samples independently drawn from the joint distribution of $(\subsetF,\subsetS)$ and let $\mathbbm{1}_{A}(\cdot)$ denote the indicator function for set $A$.
    If
    \begin{align}\label{eq:cor_cond}
         \frac{1 - \Pr\left\lbrace \hat{\mathbf{f}}(\mathbf{s}_n) \neq \mathbf{f}_n \right\rbrace}{1 - \Pr\left\lbrace \bar{\mathbf{f}} \neq \mathbf{f}_n \right\rbrace} &= \lim\limits_{N\to\infty}  \frac{\frac{1}{N}\sum_{n=1}^N \mathbbm{1}_{\lbrace\mathbf{f}_n\rbrace}(\hat{\mathbf{f}}(\mathbf{s}_n))}{\frac{1}{N}\sum_{n=1}^N \mathbbm{1}_{\lbrace\mathbf{f}_n\rbrace}(\bar{\mathbf{f}})}\\\nonumber &\equiv \lim\limits_{N\to\infty} \frac{A_N(\hat{\mathbf{f}})}{A_N(\bar{\mathbf{f}})} > 1,
    \end{align}
    $\subsetF$ is a modulator of $\setS$.
\end{corollary}

$A_N(\hat{\mathbf{f}})$ as defined in the corollary denotes the {\em accuracy} of $\hat{\mathbf{f}}$ evaluated over $N$ samples.

Since for practical applications only finitely many samples $N$ and potentially suboptimal estimators, but not $\hat{\mathbf{f}}$, are available, we close this section by extending the strict notion of \ac{GBA} modulation from Def.~\ref{def:gba_modulator} with a relaxation of \eqref{eq:cor_cond}.

\begin{definition}[Empirical GBA modulator]\label{def:empirical_modulator}
    Let $\tilde{\mathbf{f}}$ denote any estimator of $\subsetF$.
    If $\frac{A_N(\tilde{\mathbf{f}})}{A_N(\bar{\mathbf{f}})} > 1$ for $N$ samples randomly drawn from the joint distribution of $(\subsetF,\subsetS)$, $\subsetF$ is an {\em empirical} modulator {\em (of degree $N$)} of $\setS$.
\end{definition}
%
\section{Analysis}\label{sec:random_forest_model}
To overcome the problem that the optimal estimator $\hat{\mathbf{f}}$ is unknown in general, we use machine learning to train a suboptimal classifier on solving \eqref{eq:system_model:inference} approximately.
Specifically, for discrete \acp{RV} $F_j$ as assumed above, \eqref{eq:system_model:inference} constitutes a {\em classification problem} and we use an \ac{RF} model trained on a finite set of data samples to solve it.
\Acp{RF} harness the collective intelligence of multiple decision trees for classification and prediction.
Unlike other machine learning tools, \acp{RF} require minimal parameter tuning and are less prone to overfitting \cite{breiman2001random}.
This makes \acp{RF} well suited for the small data sets used in this work.

We note that due to its potential suboptimality, the estimation performance achieved with the \ac{RF}-based classifier constitutes a lower bound for the performance of $\hat{\mathbf{f}}$, if the considered data samples are representative of the true joint distribution of $(\subsetF,\subsetS)$.
To ensure that the latter condition is fulfilled, we restrict the analysis to low-dimensional subsets $\subsetF \subset \setF$ and $\subsetS \subset \setS$ avoiding the curse of dimensionality \cite{bellman2021dynamic}.
To make this notion precise, we denote the \ac{RF}-based estimator trained on any subset of bacterial species $\subsetS$ as $\hat{\mathbf{f}}_{\mathbf{S}}^{\mathrm{RF}}(\mathbf{s})$ and define the optimal subset of features for \ac{RF}-based classification as follows
{\setlength{\abovedisplayskip}{1pt}
\setlength{\belowdisplayskip}{1pt}
\begin{equation}
    \mathbf{S}^{*} = \argmax\limits_{\mathbf{S}' \subset \setS} \mathbb{E}_{\mathcal{D}}\, A_N\left(\hat{\mathbf{f}}_{\mathbf{S}'}^{\mathrm{RF}}\right),\label{eq:rf:opt_subset}
\end{equation}}
where the expectation is over different partitionings $(\mathcal{T}_i,\mathcal{V}_i)$, $\mathcal{T}_i \cup \mathcal{V}_i = \mathcal{D}$, $\mathcal{T}_i \cap \mathcal{V}_i = \emptyset$, of the available data $\mathcal{D}$ into training sets $\mathcal{T}_i$ and test sets $\mathcal{V}_i$ with $\vert\mathcal{V}_i\vert=N$.
Exactly analogously to $\mathbf{S}^{*}$, ${\subsetXs}^*$ and ${\subsetXc}^*$ are defined as the optimal subsets of features from $\setXs$ and $\setXc$, respectively.

In the following section, we will rigorously define $\hat{\mathbf{f}}^{\mathrm{RF}}(\mathbf{s})$ and detail the strategy to determine $\mathbf{S}^*$ (${\subsetXs}^*$, ${\subsetXc}^*$) as well as the metrics used in this paper in order to assess its performance.

\subsection{Random Forest Initialization and Training}
When applying the \ac{RF} classifier to a data set $\mathcal{D}$, $M=100$ independent RF models are generated.
To this end, $\mathcal{D}$ is split into training and test sets $(\mathcal{T}_i,\mathcal{V}_i)$ according to a 3:1 ratio, where the splitting is done randomly for each RF. 
Since the splitting is independent of $\mathbf{F}$, it can potentially lead to unbalanced classes in  $\mathcal{T}_i$ and/or $\mathcal{V}_i$ due to the typically small size of $\mathcal{D}$.
Every factor $F_j \in \subsetF$ corresponds to an individual class in the \ac{RF} model.

Each \ac{RF} model consists of $T=100$ decision trees. Each tree in the forest is constructed using a random subset of $\mathcal{T}_i$.
This method introduces diversity among the trees, which is critical for the model to generalize. 
Further randomness is introduced by selecting random subsets of maximum size $k=\sqrt{\vert\subsetS\vert}$ from $\subsetS$ (or, analogously, of maximum size $k=\sqrt{\vert\subsetXs\vert}$ and $k=\sqrt{\vert\subsetXc\vert}$ from $\subsetXs$ and $\subsetXc$, respectively) for each node split during tree construction.
This feature selection is performed according to a splitting criterion called Gini impurity \cite{probst2019hyperparameters}.
The Gini impurity is a popular choice due to its efficiency in measuring the disorder at a node, which helps in deciding which features are most favorable \ac{wrt} the final prediction accuracy. 
During \ac{RF} training, the splitting at each decision tree is optimized.
The \ac{RF} implementation used to generate the results in this paper is based on the Python scikit-learn package \cite{python_scikit-learn}.

\subsection{Performance Metrics}

In addition to the accuracy $A_N$ as defined in \Section{sec:system_model}, we use the F1 score as additional performance metric.
The F1 score balances recall $\mathrm{R} = \frac{\mathrm{TP}}{\mathrm{TP} + \mathrm{FN}}$ and precision $\mathrm{P} = \frac{\mathrm{TP}}{\mathrm{TP} + \mathrm{FP}}$ and is commonly used for assessing binary classification performance when datasets are unbalanced.
Here, TP and FP denote true and false positives, respectively, and TN and FN denote true and false negatives, respectively.
More specifically, F1 corresponds to the harmonic mean of precision and recall.
Formally, the F1 score of any classifier $\tilde{\mathbf{f}}$ for test set $\mathcal{V}_i$ is defined as \cite{erickson2021magician}
{\setlength{\abovedisplayskip}{1pt}
\setlength{\belowdisplayskip}{1pt}
\begin{equation}
     \mathrm{F}1(\tilde{\mathbf{f}}, \mathcal{V}_i) = 2 \frac{ \mathrm{P} \cdot\mathrm{R}}{\mathrm{P}+\mathrm{R}} = \frac{2 \mathrm{TP}}{2 \mathrm{TP} + \mathrm{FP} + \mathrm{FN}}\,,
\end{equation}}
where P, R, TP, FP, TN, and FN are evaluated over all samples in $\mathcal{V}_i$.
We note that $\mathrm{F}1(\tilde{\mathbf{f}}, \mathcal{V}_i)/\mathrm{F}1(\bar{\mathbf{f}}, \mathcal{V}_i) > 1$ implies $A_{\vert\mathcal{V}_i\vert}(\tilde{\mathbf{f}})/A_{\vert\mathcal{V}_i\vert}(\bar{\mathbf{f}})>1$, so the $\mathrm{F}1$ score can be used instead of $A_N$ to verify the condition in Def.~\ref{def:empirical_modulator} (for finite $N$).

\subsection{Method for Selecting Important Features}\label{sec:rf:feature_selection}

After \ac{RF} models have been trained on the full sets $\setS$, $\setXs$, $\setXc$, we apply the permutation importance method \cite[Chap. 10]{breiman2001random} to solve \eqref{eq:rf:opt_subset}, i.e., to find ${\subsetS}^*$, ${\subsetXs}^*$, and ${\subsetXc}^*$.
To ensure that the importance ratings are reliable, we perform 10 random permutation tests for \textit{each feature} for \textit{each RF} model on the entire data set $\mathcal{D}$.
The most important features ${\subsetS}^*$, ${\subsetXs}^*$, and ${\subsetXc}^*$ are then selected as those features whose random permutation leads to the lowest mean accuracy.
%
\section{Data Sources}\label{sec:data_sources}
\begin{table*}[!tbp]
\caption{Data Sets}
\vspace*{-0.4cm}
\begin{center}
{\def\arraystretch{1.3}\tabcolsep=2pt
 \begin{tabular}{|l | c | c | c | c |c | p{4.6cm} | p{2.5cm} | c | c|}
   \hline
 Type & Status & Sample Origin & $|\mathcal{S}|$ & $|\mathcal{X}^{\mathrm{c}}|$ & $|\mathcal{X}^{\mathrm{s}}|$ & $\mathcal{F}$ & Sample size $\vert\mathcal{D}\vert$ & Index & Reference \\ 
 \hline\hline
 Bacteria & Epilepsy & Fecal/Colon$^{(\textnormal{a})}$  & 33$^{(\textnormal{b})}$ & - & - & \{\textcolor{red}{ND},\textcolor{green}{KD}\} \newline$\times^{(\textnormal{c})}$ \{D0,D4,D8,D14\}$^{(\textnormal{d})}$ & \textcolor{red}{3/3/3/3}/\textcolor{green}{3/3/3/3} & DS-1 & \cite[Suppl. Table 2]{olson2018gut} \\
 Metabolites & Healthy & Fecal & - & 26 & - & \{KD,ND,LCD\}$^{(\textnormal{e})}$ & 10/10/10 & DS-2 & \cite[Suppl. Table 1]{kong2021ketogenic} \\
 Metabolites & Healthy & Intestinal & - & 903 & - & \{ND,HFD,KD,HK\}$^{(\textnormal{f})}$ & 6/6/6/6 & DS-3 & \cite[Suppl. Table 4]{qiu2024metagenomic} \\
 Metabolites & Cancer & Blood Plasma & - & - & 75 & \{\textcolor{red}{A375},\textcolor{green}{WM47},\textcolor{orange}{WM3311},\textcolor{blue}{WM3000}\}$^{(\textnormal{g})}$ \newline$\times$ \{ND, KD1+KD2\}$^{(\textnormal{h})}$ & \textcolor{red}{13/11+10}/\textcolor{green}{10/7+7}/\newline\textcolor{orange}{10/11+10}/\textcolor{blue}{12/12+12} & DS-4 & \cite[Suppl. Table 2]{weber2022ketogenic} \\

 \hline
 \end{tabular}
 }
 \end{center}
\begin{tablenotes}[flushleft]\footnotesize\setlength\itemsep{-0.0cm}
\setlength{\columnsep}{0.8cm}
\setlength{\multicolsep}{0cm}
  \begin{multicols}{2}
\item[] (a): There is no distinction between colon and fecal samples in the data set.
\item[] (b): Taxonomy level: Bacterial family level.
\item[] (c): $\times$ denotes the Cartesian product.
\item[] (d): Day 0 (D0), Day 4 (D4), Day 8 (D8), Day 14 (D14). 
\item[] (e): Low-carbohydrate diet (LCD), normal diet (ND).
\item[] (f): High-fat diet (HFD), high-fat to ketogenic conversion (HK).
\item[] (g): Human melanoma cell lines A375, WM47, WM3311, and WM3000.
\item[] (h): Long‑chain triglyceride‑based ketogenic diet (KD1) and Long‑chain triglyceride‑based ketogenic diet supplemented with C8 and C10 medium chain triglycerides (KD2) aggregated as $\text{KD}=\text{KD1}+\text{KD2}$.
  \end{multicols}
\end{tablenotes}

 \label{Table:data_sets}
 \vspace*{-0.4cm}
\end{table*}
The proposed machine learning model is tested on four different peer-reviewed mouse datasets that are available online \cite{kong2021ketogenic,qiu2024metagenomic,weber2022ketogenic,olson2018gut}.
All datasets study the impact of \ac{KD} on either $\setS$, $\setXs$,  or $\setXc$.
Table~\ref{Table:data_sets} lists the used datasets.
Since the health status of the host is a potential GBA modulator, datasets from healthy (DS-2 \cite{kong2021ketogenic} and DS-3 \cite{qiu2024metagenomic}) as well as from disease mouse models (DS-1 \cite{olson2018gut} and DS-4 \cite{weber2022ketogenic}) are included.
Specifically, the mice in \cite{olson2018gut} serve as animal models for \ac{DRE} and a mouse model of human melanoma (skin cancer) is considered in \cite{weber2022ketogenic}.
Note that the sample sizes as reported in Table~\ref{Table:data_sets} are relatively small, ranging from $\vert\mathcal{D}\vert=24$ to $\vert\mathcal{D}\vert=125$, since they are obtained from animal models.
%
\section{Evaluation}\label{sec:results}
\subsection{Impact of KD on the Gut Microbiota in a Mouse Model of Epilepsy}

\begin{figure}
    \centering
    \includegraphics[width=0.99\linewidth]{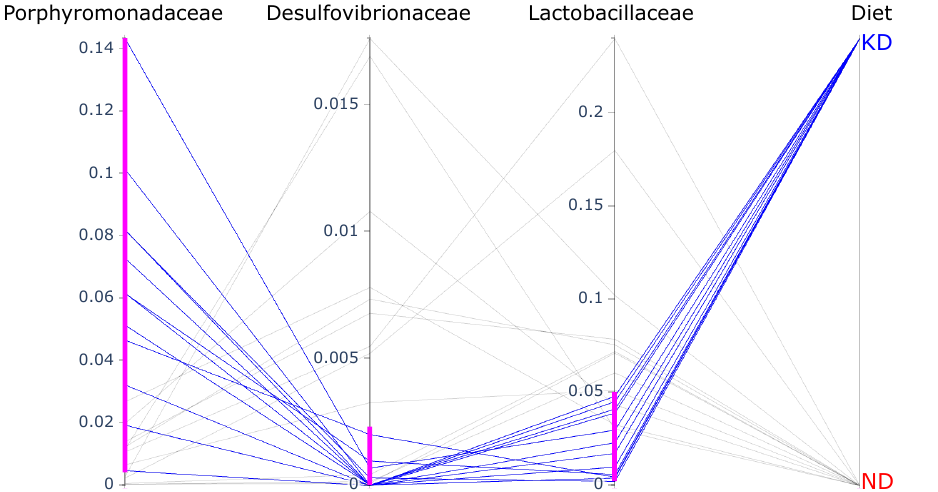}
    \vspace*{-3mm}
    \caption{\small Relative abundances of the three most important bacterial species $\subsetS^*$ for discriminating \ac{KD} and \ac{ND} in the dataset from \cite{olson2018gut}. The three left-most vertical coordinate axes correspond to the three bacterial species in $\subsetS^*$, the value of $F_1$, \ac{KD} or \ac{ND}, is displayed on the right. Data samples are illustrated as blue and gray lines that interconnect the different coordinate axes. Blue lines correspond to data samples selected according to the intervals highlighted in pink.}
    \vspace*{-5mm}
    \label{fig:ds7_par_coord}
\end{figure}

First, we study whether \ac{KD} modulates $\setS$ in the mouse model of epilepsy.
To this end, we consider the dataset from \cite{olson2018gut} (DS-1 in Table~\ref{Table:data_sets}) and train the \ac{RF} model on predicting $\subsetF = \lbrace F_1 \rbrace$, where $F_1 \in \lbrace \mathrm{KD}, \mathrm{ND} \rbrace$, and $F_1 = \mathrm{KD}$ for data samples from mice fed with \ac{KD} and $F_1 = \mathrm{ND}$ for mice fed with \ac{ND}. 
Applying the feature selection strategy detailed in \ref{sec:rf:feature_selection}, we obtain $\subsetS^*=\lbrace \text{Lactobacillaceae},\allowbreak \text{Desulfovibrionaceae},\allowbreak \text{Porphyromonadaceae} \rbrace$, which is in agreement with previous studies \cite{alexander2024diet-dependent, paoli2019ketogenic}.
Fig.~\ref{fig:ds7_par_coord} illustrates how the features in ${\subsetS}^*$ are used to successfully classify the data samples into \ac{KD} and \ac{ND}.
Specifically, it is evident from Fig.~\ref{fig:ds7_par_coord} that the mice from DS-1 can be perfectly clustered into \ac{KD} and \ac{ND} mice according to the relative abundances of the gut bacteria in ${\subsetS}^*$
The F1 score of the resulting \ac{RF} model is $0.939053$, exceeding by far the F1 score of $0.\bar{6}$ that is achieved with a model that classifies according to the majority class (either \ac{KD} or \ac{ND} in this case), i.e., $\bar{\mathbf{f}}$, cf.~\eqref{eq:system_model:priori_estimator}.
We conclude from Def.~\ref{def:empirical_modulator} that \ac{KD} is an empirical \ac{GBA} modulator.
Since the diet is an externally controllable variable, \ac{KD} is also a feasible, empirical \ac{GBA} modulator for the epileptic mice from the dataset from \cite{olson2018gut}.

\subsection{Metabolomic Fingerprint of KD Compared to Other High-Fat, Low-Carbohydrate Diets in Healthy Mice}

Next, we ask whether \ac{KD} is distinguishable from other dietary interventions in terms of $\setXc$.
To this end, we consider the datasets from \cite{kong2021ketogenic} and \cite{qiu2024metagenomic} (DS-2 and DS-3, respectively, in Table~\ref{Table:data_sets}).
Specifically, \ac{KD}, \ac{LCD}, and \ac{ND} are considered as dietary interventions in \cite{kong2021ketogenic}, while \ac{KD}, \ac{HFD}, \ac{HK}, and \ac{ND} are considered in \cite{qiu2024metagenomic}.
The metabolic profiles from \cite{kong2021ketogenic} were obtained with a {\em targeted} assay, i.e., only the concentrations of $\vert\setXc\vert=26$ preselected metabolites were measured.
In contrast, the data from \cite{qiu2024metagenomic} was obtained in an {\em untargeted} manner, i.e., without preselecting specific metabolites, and comprises data for $\vert\setXc\vert=903$ different metabolites.

\begin{figure}
    \centering
    \includegraphics[width=0.94\linewidth]{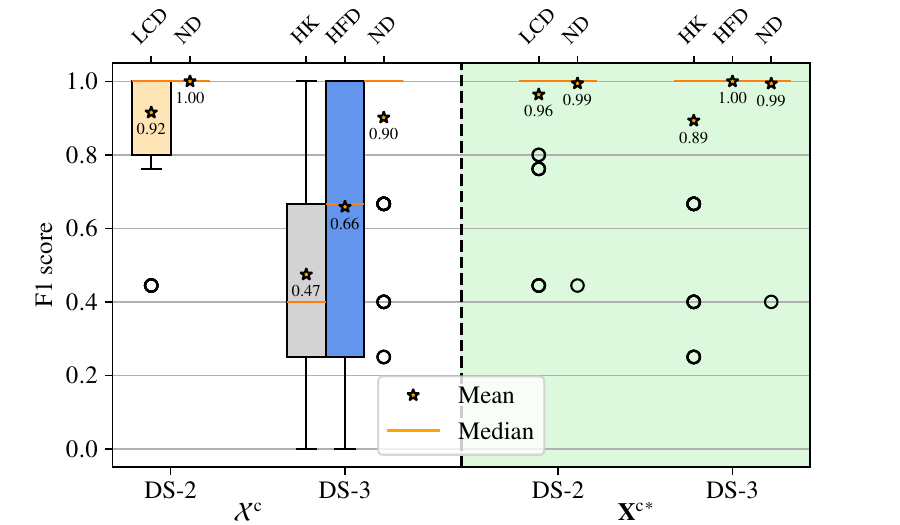}
    \vspace*{-3.5mm}
    \caption{\small F1 scores for the different binary classification tasks \ac{KD} vs.~\{\ac{ND},\ac{LCD},\ac{HFD},\ac{HK}\}. The boxplots depict the distributions of F1 scores across different partitionings of the datasets into training and test sets (the ratio of training samples to test samples remains fixed) for the complete set of colon metabolites $\setXc$ (left, white background) and for the optimal subset of metabolites ${\subsetXc}^*$ (right, green background).}
    \vspace*{-6mm}
    \label{fig:f1_ds4_5}
\end{figure}

Fig.~\ref{fig:f1_ds4_5} illustrates the F1 scores obtained for these datasets for $M=100$ different partitionings $(\mathcal{T}_i,\mathcal{V}_i)$ on the full set of colon metabolites $\setXc$ as well as on the set of most important metabolites ${\subsetXc}^*$.
We observe from Fig.~\ref{fig:f1_ds4_5} that for DS-2 the \ac{RF} classifier can distinguish very well between \ac{KD} and \ac{ND} as well as between \ac{KD} and \ac{LCD} based on both $\setXc$ and ${\subsetXc}^*$.
Here, distinguishing \ac{LCD} and \ac{KD} is the more challenging task, since the \ac{LCD} is more similar to the \ac{KD} in nutrient composition as compared to \ac{ND}.
However, the performance of the \ac{RF} model for DS-3 on $\setXc$ is acceptable only for the simplest task, \ac{KD} vs.~\ac{ND}.
\ac{HK} and \ac{HFD} cannot be distinguished reliably based on $\setXc$.
This outcome is expected, since \ac{HK} and \ac{HFD} are much similar to \ac{KD} \ac{wrt} their respective nutrient composition as compared to \ac{ND}.
Finally, we observe from Fig.~\ref{fig:f1_ds4_5} that all different diets can perfectly be distinguished from \ac{KD} based on the important factors ${\subsetXc}^*$ with median $\mathrm{F}1$ scores identical to 1.
This result has two important consequences.
First, the \ac{KD} modulates $\setXc$ according to Defs.~\ref{def:gba_modulator} and \ref{def:empirical_modulator}.
Second, the important factor selection strategy adopted in this paper can effectively compensate the curse of dimensionality that otherwise arises in the high-dimensional data space of DS-3.

\subsection{Metabolomic Fingerprint of KD in a Mouse Model of Cancer}

Finally, we study the impact of the \ac{KD} on $\setXs$ in a mouse model of human melanoma \cite{weber2022ketogenic} (DS-4 in Table~\ref{Table:data_sets}).
Specifically, we ask two questions, namely (i) whether \ac{KD} and/or cancer type (A375, WM47, WM3311, WM3000) are (empirical) modulators of $\setXs$ and (ii) whether the impact of \ac{KD} on $\setXs$ is the same irrespective of the type of cancer the host has.

To answer the first question, we evaluated the $\mathrm{F}1$ scores of \ac{RF} models trained to classify the type of diet and the type of cancer, respectively.
For diet prediction, the model achieved a median $\mathrm{F}1$ score of $0.97$, and an accuracy of $>0.98$ for cancer type prediction (after reducing the feature space to ${\subsetXs}^*$).
Hence, both \ac{KD} and cancer type modulate $\setXs$ (empirically) according to Def.~\ref{def:empirical_modulator}, though from the two only \ac{KD} is a feasible modulator, cf.~Def.~\ref{def:feasible_modulator}.

To answer the second question, we trained different \ac{RF} models to classify \ac{KD} vs.~\ac{ND} for specific types of cancer.
In this way, we obtain $4$ different estimators ${\hat{\mathbf{f}}}^{\text{RF}}_\text{A375}$, ${\hat{\mathbf{f}}}^{\text{RF}}_\text{WM47}$, ${\hat{\mathbf{f}}}^{\text{RF}}_\text{WM3311}$, and ${\hat{\mathbf{f}}}^{\text{RF}}_\text{WM3000}$, where each of these was trained on data samples with the cancer type indicated in the subscript and the respective optimal sets of serum metabolites ${\subsetXs}^*$ denoted in the following as ${\subsetXs}^{*}_\text{A375}$, ${\subsetXs}^{*}_\text{WM47}$, ${\subsetXs}^{*}_\text{WM3000}$, and ${\subsetXs}^{*}_\text{WM3311}$.
The performance of each of these estimators is then evaluated on test data samples with the same cancer type and on data samples with different cancer types.
Table~\ref{tab:ds6_plasma} summarizes the resulting mean $\mathrm{F}1$ scores.

First, we observe from Table~\ref{tab:ds6_plasma} that each estimator works perfectly ($\mathrm{F}1$ score identical to $1$) on data with the same cancer type as it was trained on.
Hence, we conclude that \ac{KD} is a robust GBA modulator according to Def.~\ref{def:robust_modulator}, where, here, $\subsetF_1=\lbrace \text{KD}, \text{ND} \rbrace$ and $\subsetF_2=\lbrace \text{A375}, \text{WM47}, \text{WM3311}, \text{WM3000} \rbrace$.
Second, we oberve from Table~\ref{tab:ds6_plasma} that the estimation performance of each estimator deteriorates when applied to data samples with other cancer types.
For example, ${\hat{\mathbf{f}}}^{\text{RF}}_{\text{A375}}$ achieves only a mean $\mathrm{F}1$ score of $0.75$ when applied to data of cancer type WM3000.
This observation points towards differences in how \ac{KD} modulates the \ac{GBA} depending on the cancer type.

\sisetup{round-mode=places, round-precision=2} 
\begin{table}[tp]
    \centering
    \vspace*{1mm}
    \caption{F1 scores for different train (columns) and test (rows) conditions (DS-4)}
    \vspace*{-1.4mm}
    \begin{tabular}{c|S[table-format=1.2]|S[table-format=1.2]|S[table-format=1.2]|S[table-format=1.2]} \toprule
    serum & \textbf{A375} & \textbf{WM3000} & \textbf{WM3311} & \textbf{WM47} \\ \midrule
    \textbf{A375}   & 1.000000 & 0.894659 & 0.974508 & 1.000000 \\
    \textbf{WM3000} & 0.993313 & 0.997720 & 0.929835 & 0.987460 \\
    \textbf{WM3311} & 0.752118 & 0.883044 & 0.997091 & 0.933621 \\
    \textbf{WM47}   & 1.000000 & 0.962897 & 0.763617 & 1.000000 \\ 
    \end{tabular}
    \label{tab:ds6_plasma}
    \vspace*{-1.5mm}
\end{table}

\begin{figure}
    \centering
    \includegraphics[width=0.94\linewidth]{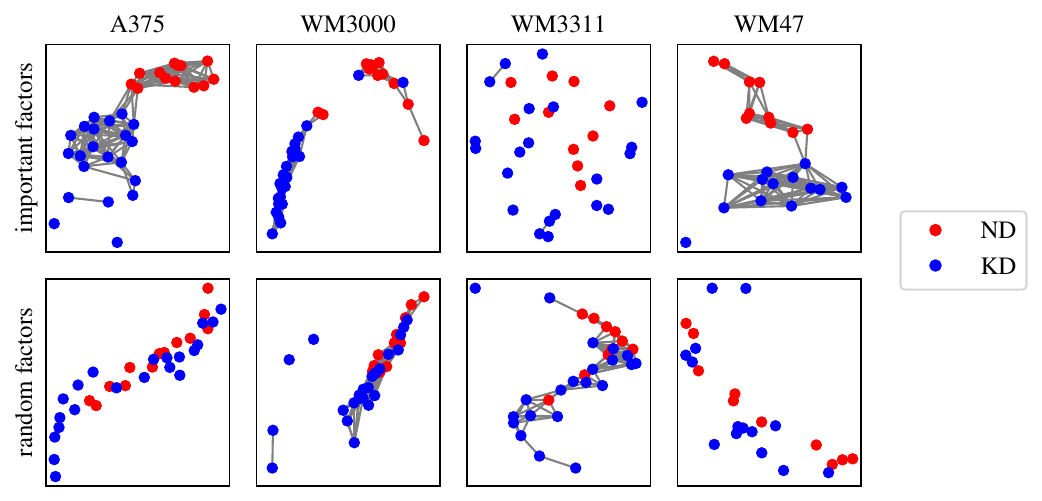}
    \vspace*{-4mm}
    \caption{\small Network plot based on the similarity of the data samples from DS-4 \ac{wrt} the important factors ${\subsetXs}^*$ (top) and a random selection of $\vert{\subsetXs}^*\vert$ (bottom) factors from $\setXs$. Blue and red dots indicate data samples with \ac{KD} and \ac{ND}, respectively.}
    \vspace*{-5.5mm}
    \label{fig:ds6_network}
\end{figure}

To further elucidate this relationship, we study ${\subsetXs}^{*}_\text{A375}$, ${\subsetXs}^{*}_\text{WM47}$, ${\subsetXs}^{*}_\text{WM3000}$, ${\subsetXs}^{*}_\text{WM3311}$ in the following.
Fig.~\ref{fig:ds6_network} illustrates how the important factors identified via \eqref{eq:rf:opt_subset} cluster the data points for each cancer type.
Specifically, we observe from Fig.~\ref{fig:ds6_network} that ${\subsetXs}^*$ provides a clearly visible clustering of the data samples for each cancer type (top panels in Fig.~\ref{fig:ds6_network}), whereas no such cluster structures are recognizable for a random selection of factors (bottom panels in Fig.~\ref{fig:ds6_network}).
Furthermore, we identified the following important factors
{\setlength{\abovedisplayskip}{1pt}
\setlength{\belowdisplayskip}{1pt}
\begin{align}
\vspace*{-3mm}
 &{\subsetXs}^{*}_\text{A375} &=& \lbrace \text{$\alpha$-Aminoadipic acid}, \text{1-Methylhistidine},\\[-1ex]\nonumber
 &&&\,\text{$\beta$-Aminobutyric acid},\text{Proline Betaine} \rbrace,\\[-0.3ex]
 &{\subsetXs}^{*}_\text{WM47} &=& \lbrace \text{$\alpha$-Aminoadipic acid}, \text{$\beta$-Aminobutyric acid},\\[-3ex]\nonumber
 &&&\,\text{EPA}, \text{1-Methylhistidine} \rbrace,\\[-0.3ex]
 &{\subsetXs}^{*}_\text{WM3000} &=& \lbrace \text{$\beta$-Aminobutyric acid}, \text{1-Methylhistidine}, \text{DHA},\\[-3ex]\nonumber
 &&&\,\text{Threonine} \rbrace,\\[-0.3ex]
 &{\subsetXs}^{*}_\text{WM3311} &=& \lbrace \text{Lysine}, \text{$\alpha$-Aminoadipic acid}, \text{Tryptophan},\\[-3ex]\nonumber
 &&&\,\text{Glutamine} \rbrace.
 \vspace*{-3mm}
\end{align}}
We conclude that some serum metabolites, such as $\alpha$-Aminoadipic acid, 1-Methylhistidine, and $\beta$-Aminobutyric acid, appear to be indicative of \ac{KD} irrespective of the cancer type and these findings are in line with what has been previously shown by others \cite{peng2025new} and in general with the effect of \ac{KD} on aminoacids metabolims \cite{yudkoff2007ketogenic}.
However, other metabolites are informative of \ac{KD} only for specific cancer types, such as Proline Betaine for A375, EPA for WM47, DHA and Threonine for WM3000, and Lysine, Tryptophan, and Glutamine for WM3311.
This finding is in line with previous evidence of \ac{KD} effect on Tryptophan metabolism \cite{effinger2023ketogenic} and aminoacids metabolisms \cite{yudkoff2007ketogenic}.

%
\vspace*{-1mm}
\section{Conclusion}\label{sec:conclusion}
\vspace*{-1mm}
In this paper, we present a novel framework for \ac{GBA-MC}.
The framework defines formal criteria for identifying potential therapeutic targets for \ac{GBA}-based treatment of neurological diseases.
To demonstrate the applicability of the general model to real-world data, we developed a machine learning approach and examined the effects of dietary interventions, particularly the \ac{KD}, on the gut microbiome and metabolic profiles of healthy and diseased mice.
Using the proposed approach, we confirmed the role of the \ac{KD} as a potential therapeutic target that modulates \ac{GBA}.
The proposed model could also identify specific bacterial species and metabolites likely to play a major role in mediating the impact of the \ac{KD} on the host.
These findings open up promising avenues for further research on the design of highly effective, personalized dietary interventions for treating neurological disorders.
%
\vspace*{-2mm}
\bibliographystyle{IEEEtran}
\bibliography{sample-base}
\end{document}